\documentclass{article} 
\usepackage{iclr2025_conference,times}

\usepackage{graphicx}
\usepackage{booktabs} 
\usepackage{setspace}

\usepackage{graphicx}

\usepackage{amsmath,amsfonts,bm}

\def\eqref#1{equation~\ref{#1}}

\def\1{\bm{1}}

\DeclareMathAlphabet{\mathsfit}{\encodingdefault}{\sfdefault}{m}{sl}
\SetMathAlphabet{\mathsfit}{bold}{\encodingdefault}{\sfdefault}{bx}{n}

\usepackage{hyperref}
\usepackage{url}
\usepackage{colortbl}
\usepackage{booktabs}
\usepackage{multirow}
\usepackage{xcolor}
\usepackage{lipsum}
\usepackage[most]{tcolorbox} 
\usepackage{ulem}

\definecolor{teagreen}{HTML}{e1f8bd}
\definecolor{lightblue}{HTML}{d3f4ff}
\definecolor{front-color}{HTML}{ffe9b8}
\definecolor{lightpink}{HTML}{ebdef0 }
\definecolor{Gray}{gray}{0.7}

\definecolor{Magenta}{rgb}{0.8, 0.1, 0.6}

\usepackage{color, soul}
\sethlcolor{yellow}
\setstcolor{red}
\setulcolor{green}

\title{Capybara-OMNI: An Efficient Paradigm for Building Omni-Modal Language Models}

\author{Xingguang Ji, \
Jiakang Wang, \
Hongzhi Zhang, \
Jingyuan Zhang, \
Haonan Zhou, \\
\textbf{Chenxi Sun}, \
\textbf{Yahui Liu}, \
\textbf{Qi Wang}, \
\textbf{Fuzheng Zhang}
\\ 
Kuaishou Technology
}

\iclrfinalcopy 
\begin{document}

\maketitle

\begin{abstract}
With the development of Multimodal Large Language Models (MLLMs), numerous outstanding accomplishments have emerged within the open-source community. 
Due to the complexity of creating and training multimodal data pairs, it is still a computational and time-consuming process to build powerful MLLMs.  
In this work, we introduce Capybara-OMNI\footnote{We named our model Capybara-OMNI, for we hope that it can provide assistance to those who are in need.
}, an MLLM that trains in a lightweight and efficient manner and supports understanding text, image, video, and audio modalities.
We present in detail the framework design, the data construction, and the training recipe, to develop an MLLM step-by-step to obtain competitive performance. 
We also provide exclusive benchmark utilized in our experiments to show how to properly verify understanding capabilities across different modalities.
Results show that by following our guidance, we can efficiently build an MLLM that achieves competitive performance among models of the same scale on various multimodal benchmarks. 
Additionally, to enhance the multimodal instruction-following and conversational capabilities of the model, we further discuss how to train the chat version upon an MLLM understanding model, which is more in line with user habits for tasks like real-time interaction with humans. 
We publicly disclose the Capybara-OMNI model, along with its chat-based version. The disclosure includes both the model weights, a portion of the training data, and the inference codes, which are made available on GitHub~\footnote{The source codes and data are available at:\url{https://github.com/stoney0062/CAPYBARA-OMNI}.}.

\end{abstract}

\section{Introduction}
\label{sec:introduction}

Recently, Large Language Models (LLMs) have witnessed rapid development~\citep{gpt4,anthropic2024claude35,team2023gemini,qwen2,deepseekai2024deepseekv3technicalreport}. In particular, the integration of traditional multimodal understanding with LLMs has enabled these models to be applied to the fields of images, videos, and audio, termed as Multimodal Large Language Models (MLLMs). 
For example, the emergence of GPT-4o~\citep{openai2024gpt4o} has demonstrated its powerful multimodal understanding capabilities and real-time interaction features, leading a new direction in human-computer interaction.

Until now, there are numerous explorations on improving the multimodal-understanding capabilities for MLLMs. From the perspective of the number of modalities,
Some of these models are solely focused on understanding single-modality information. For example, InternVL2~\citep{chen2023internvl}, LLaVA~\citep{liu2023llava}, LLaVA-OneVision~\citep{li2024llavaonevisioneasyvisualtask} and MiniCPM~\citep{yao2024minicpmvgpt4vlevelmllm} concentrate on visual information understanding, and Qwen2-Audio~\citep{Qwen2-Audio} specializes in audio-modality information understanding.
Meanwhile, some models manage to explore understanding full-modality information, such as VITA~\citep{fu2025vita15gpt4olevelrealtime} and MINI-OMNI~\citep{xie2024mini}. However, due to the complexity of full-modality information, improving the understanding capabilities across various modalities remains a challenging yet crucial problem. 

\begin{figure}[ht]
\centering
\includegraphics[width=\textwidth]{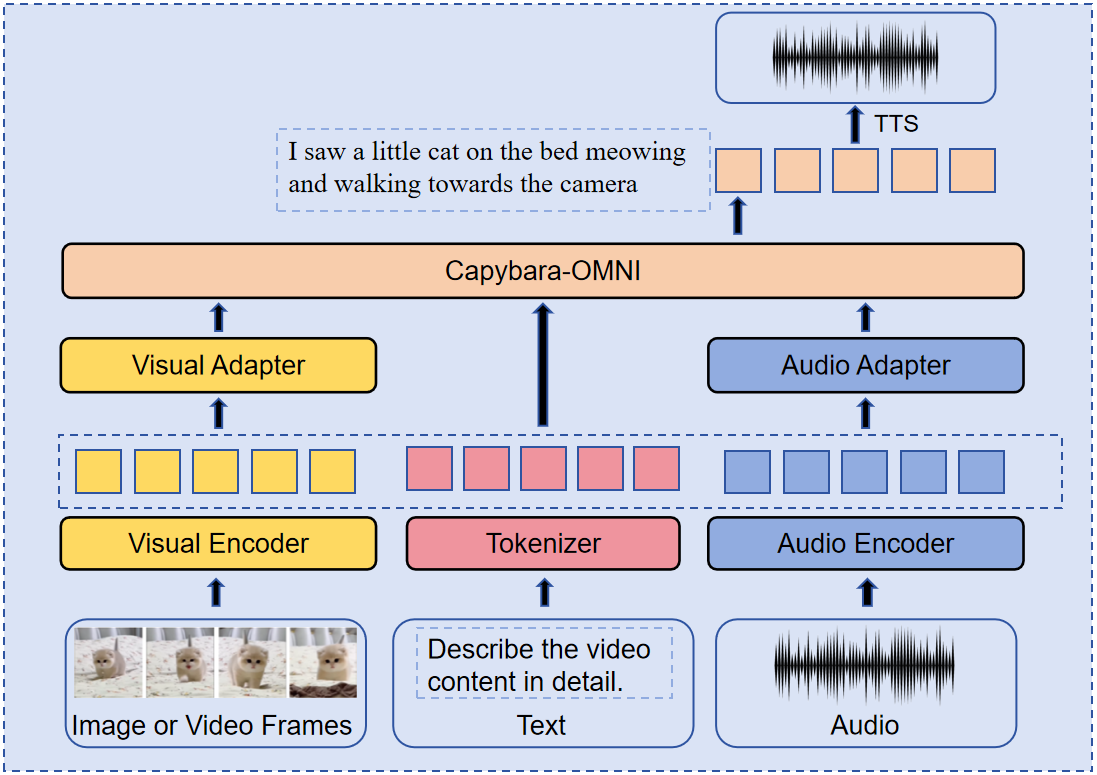}
\caption{The architecture of Capybara-OMNI. It supports various modalities as input, including text, image, video and audio. Moreover, it expands the output modalities to text and audio. 
}
\label{fig:1}
\end{figure}

On one hand, there are significant differences among various modalities. For example, 
visual data (\textit{e.g.}, images and videos) usually presents the spatial/temporal continuous information with a 2D or 3D grid shape, and audio data perceives over time as a sequence of pressure waves that presents the temporal continuous information. 
Hence, it requires a plugin encoder to efficiently extract meaningful and useful features for each modality in a MLLM.

On the other hand, the information density in each modality varies a lot. 
For example, we can use several words (\textit{e.g.}, ``a white cat" ) to represent a cat, while we need to use an image with dense pixels (\textit{e.g.}, image size 224$\times$224$\times$3) to describe the cat, and a ``meow" voice to indicate the existence of a cat. 
It is common that the information can be interfered in understanding cross modalities.
We observe that once the audio understanding ability is introduced, the visual understanding ability of the model may decline substantially without a crafted training design.

To address these challenges, we propose an ingenious design of the training process and several data optimization techniques. 
In this manner, we build an MLLM endowed with full-modality understanding capabilities, named Capybara-OMNI, through a training procedure of low complexity and resource efficiency, as shown in Figure~\ref{fig:1}. 
For example, it takes around 1.5M training data and 5 hours to efficiently expand a Vision-Language Model (VLM) with audio understanding capabilities in our experiments.

In summary, the contributions of this paper are as follows:
\begin{enumerate}
    \item We show a detailed roadmap to endow a LLM  
    with competitive visual and audio understanding capabilities through lightweight training while addressing the issue of mutual interference among modal understanding capabilities.
    \item We develop Capybara-OMNI, which is a model that can handle text, image, video, and audio input, based on our proposed lightweight training process, demonstrating strong understanding capabilities across all modalities. 
    \item We propose an efficient method for enhancing the instruction-following and interaction capabilities. With meticulously constructed multimodal instruction fine-tuning data, the fine-tuned Capybara-OMNI shows strong human-computer interaction capabilities and the ability to understand complex multimodal scenarios.
\end{enumerate}

\section{Related Works}
\label{sec:related-work}

\paragraph{Vision Language Models}
Recently, Vision-Language Models (VLMs)~\citep{alayrac2022flamingo,chen2023internvl,liu2023llava,openai2023gpt4v} have experienced rapid development. LLaVA-OneVision~\citep{li2024llavaonevisioneasyvisualtask} is the first single model that can simultaneously break through the performance boundaries of open multimodal models in three important computer vision scenarios: single-image, multi-image, and video scenarios. MiniCPM-V~\citep{yao2024minicpmvgpt4vlevelmllm} achieves efficient compression of image tokens while demonstrating strong capabilities in visual understanding. Qwen2-VL~\citep{wang2024qwen2vlenhancingvisionlanguagemodels} supports the understanding and analysis of long videos and also has advanced interaction methods. 

\paragraph{Audio Language Models}
As a direct mode of communication in human-computer interaction, audio LLM has also received extensive research attention. Qwen-Audio~\citep{Qwen2-Audio} possesses strong audio understanding and instruction-following capabilities, and is capable of intelligently comprehending the content within audio and responding appropriately by following voice commands. GLM-4-Voice~\citep{zeng2024glm4voiceintelligenthumanlikeendtoend} is an end-to-end large speech model that can directly understand and generate Chinese and English speech and conduct real-time voice conversations. Moshi~\citep{défossez2024moshispeechtextfoundationmodel}, the first streaming, full-duplex speech dialogue model, significantly enhances linguistic quality through the parallel generation of text and audio. Mini-Omni~\citep{xie2024mini} adopts a similar approach.

\paragraph{Multi-modal Interaction Models}

GPT-4o~\citep{hurst2024gpt} represents a milestone in the development of MLLMs, offering robust multimodal understanding and end-to-end real-time conversational capabilities. Recently, several excellent open-source models~\citep{fu2025vita15gpt4olevelrealtime,xie2024mini,li2024baichuanomni} have emerged that possess functionalities similar to GPT-4o. These models are capable of processing and analyzing video, image, text, and audio modalities simultaneously, and offer advanced multimodal interaction experiences. 
\begin{figure}[ht]
\centering
\includegraphics[width=1\textwidth]{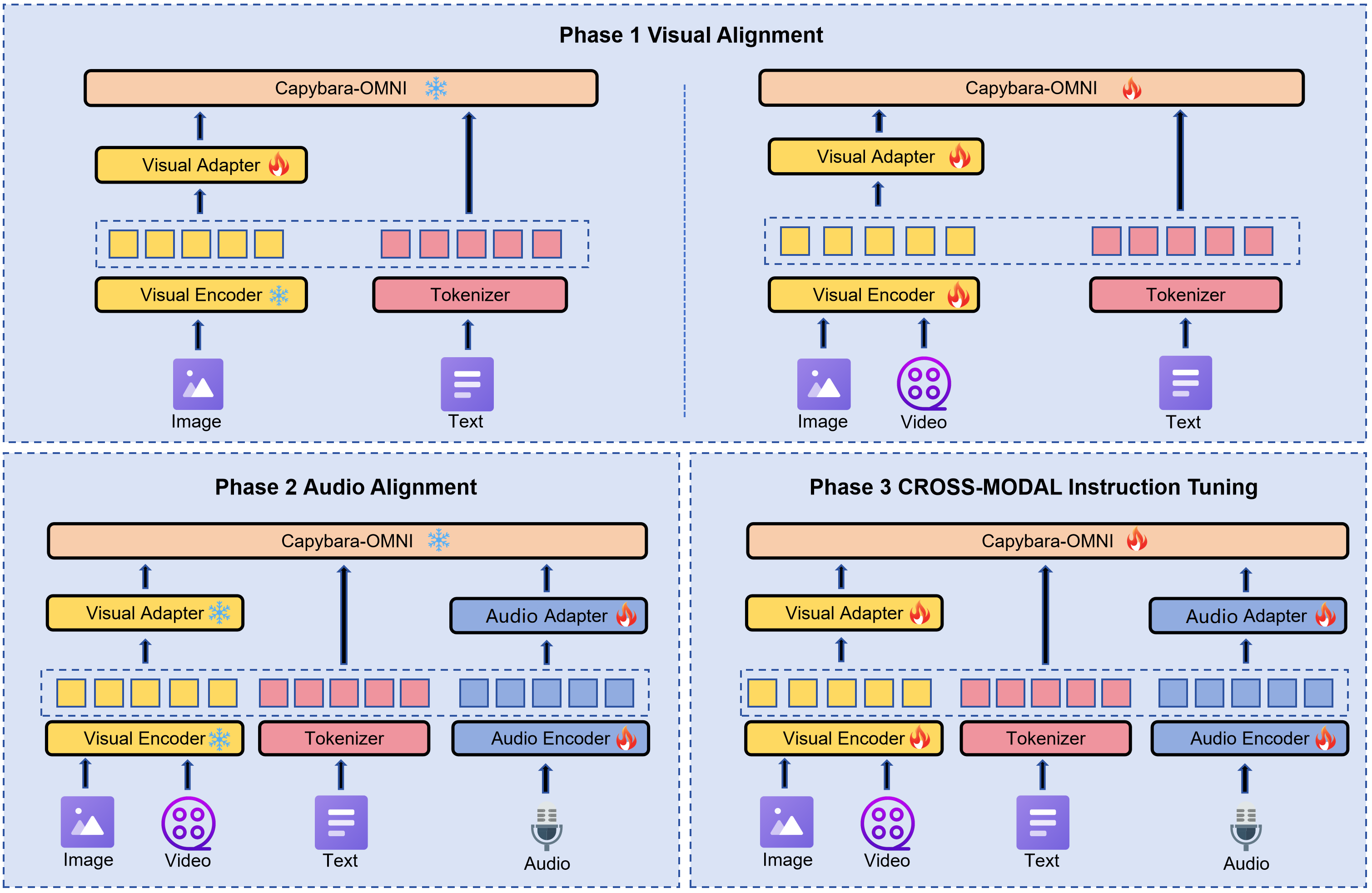}
\caption{The overview of training Capybara-OMNI. We propose to handle multiple modalities in a progressive manner, which consists of three main stages. We start from the alignment between image and text and then expand to other modalities. In each stage, we freeze (\textit{i.e.}, the blue snow symbol) or unfreeze (\textit{i.e.}, the red fire symbol) the module parameters to update the model.}
\label{fig:2}
\end{figure}

\section{Training Design}
\label{sec:training}

In light of the distinct input data modalities, the training process is partitioned into three phrases: visual alignment, audio alignment, and cross-modal instruction tuning, as shown in Figure~\ref{fig:2}. In each stage, we utilize corresponding training data and update some or entire model parameters in a step-by-step manner. In this chapter, the first focus on the introduction of the initial two stages (\textit{i.e.}, visual alignment and audio alignment). These two phrases endow the MLLM model with the capacity to comprehend visual and audio information, respectively.

\vspace{-1em}
\subsection{Visual Alignment}
\label{subsec:visual_align}
\vspace{-1em}

\paragraph{Architecture}
We build our omnimodal understanding model upon the architecture of Capybara-VL-7B~\citep{zhang2024capybaravl}, which is a vision-language model. The visual module of our proposed Capybara-OMNI consists of three components: a visual encoder, a MLP adaptor, and a large language model (LLM), inspired by the structure of LLaVA~\citep{liu2023llava}.

Specifically, we utilize the SigLIP~\citep{zhai2023sigmoid} as our vision encoder, excluding its final hidden layer. Drawing inspiration from LLaVA-UHD~\citep{guo2024llava}, we interpolate the position embeddings of the ViT by following NaViT~\citep{dehghani2023patchnpacknavit} to accommodate variable aspect ratios and divide high-resolution images into subimages. To determine the optimal number of splits and the splitting methodology for each input image, a scoring function from MiniCPM~\citep{yao2024minicpmvgpt4vlevelmllm} is used. The original image is resized to match the target size while maintaining its aspect ratio. For high-resolution images, a thumbnail is generated to provide a global overview, whereas the image is segmented into multiple sub-images to capture finer details. In simpler cases where splitting is unnecessary, the image is utilized in its entirety.

To connect the vision encoder with the LLM, we employ a MLP, a two-layer perceptron with layer normalization, to transform visual features into the input space of the LLM. Initially, these visual embeddings are reshaped into a 2D shape and then compressed by using 2$\times$2 bilinear interpolation, effectively reducing image tokens from 1024 to 256. The compressed features are  flattened in a left-to-right, top-to-bottom sequential order. To indicate image row breaks, a special token is appended at the end of each row. This design facilitates the model's ability to reconstruct the original 2D layout from the flattened sequence. Finally, these visual embeddings are concatenated with text embeddings to form the input for the LLM. For the LLM, we utilize Qwen2.5-7B~\citep{yang2024qwen2}, which excels in understanding, reasoning, and generating coherent textual output. 

\begin{figure}[ht]
\centering
\includegraphics[width=\textwidth]{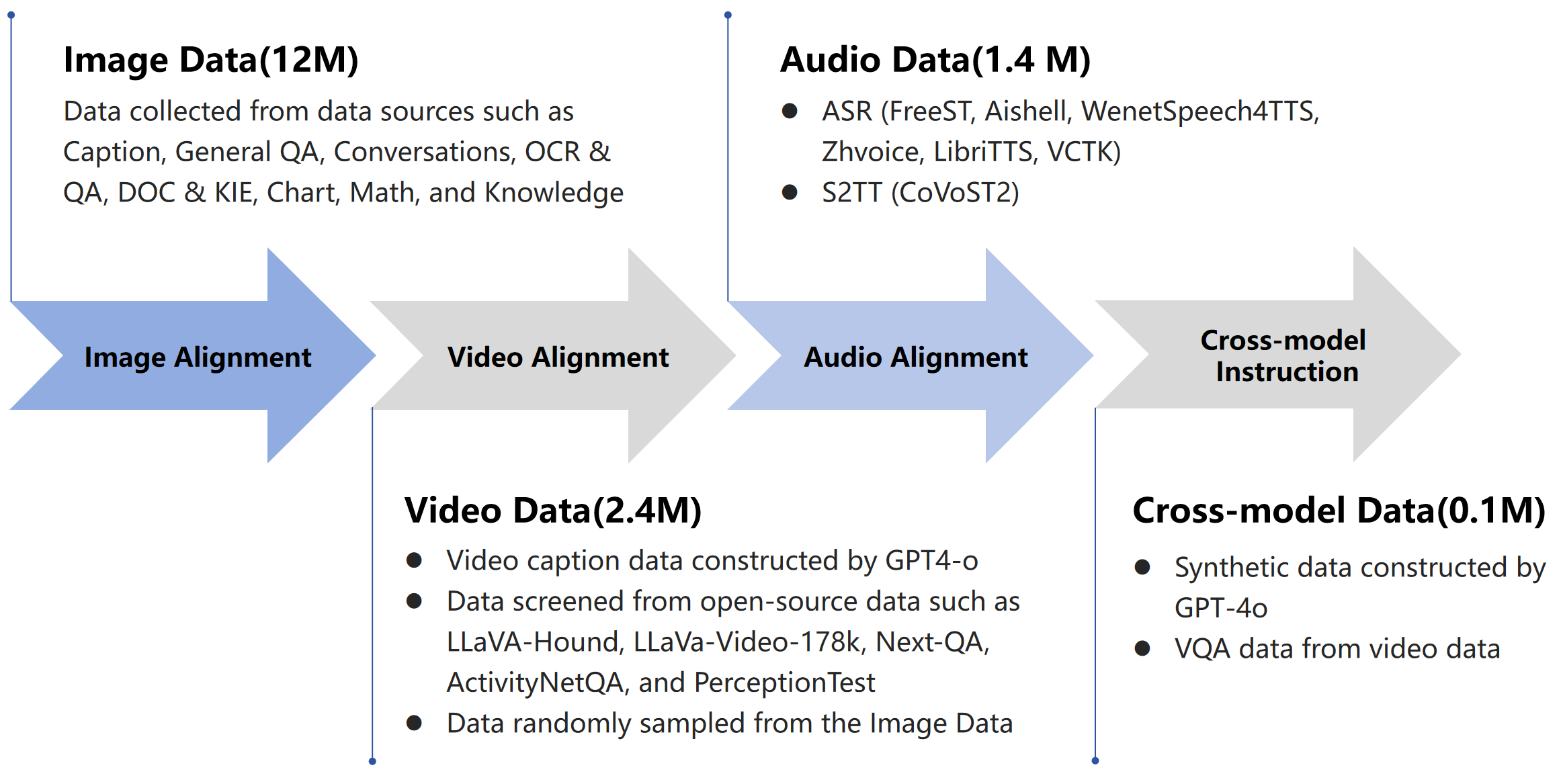}
\caption{The training data construction process for Capybara-OMNI.}
\label{fig:3}
\end{figure}

\paragraph{Training Data}
The construction process of the training data is shown in Figure~\ref{fig:3}. In terms of image training data, we collect more than 38 million samples from more than 200 different sources. After steps such as removing duplicate samples, filtering low-quality responses, and filtering pure text questions, we collect approximate 12 million high-quality samples. In addition, we use the Qwen2-VL-72B and InternVL-76B models to generate detailed answers by using methods such as prompt isolation, rewriting of answers, and generating Chain-of-Thought (CoT) answers. We automatically remove the low quality answers according to the evaluation results of GPT-4o. 
See detailed data sources in Table~\ref{tab:6} in the Appendix.

In the video understanding stage, we collect a total of 2.4 million training data. High-quality videos are collected from open-source video websites, and corresponding video captions are constructed using the GPT-4o. Moreover, some data was collected and filtered from open-source datasets such as LLaVA-Hound~\citep{zhang2024directpreferenceoptimizationvideo}, LLaVA-Video-178k~\citep{zhang2024videoinstructiontuningsynthetic}, Next-QA~\citep{xiao2021next}, Activitynet-QA~\citep{yu2019activityqa} and PerceptionTest~\citep{patraucean2023perception}. In addition, 600K data samples are uniformly sampled from of the image alignment to preserve the image understanding capability of the model.

\paragraph{Training Process}

There are three sub-stages in the image understanding stage. In the first sub-stage, we focus on high-level alignment between the vision encoder and the LLM using coarse-grained caption data, with the LLM and vision encoder parameters frozen and updating only the adapter. The second sub-stage enhances fine-grained visual concept alignment and high-resolution handling by splitting input images into up to four sub-images for better detail capture, using high-resolution fine-grained captions and text-rich data for training, and unfreezing all model parameters for adaptation. In the final sub-stage, we improve the multi-task and dialogue capabilities by training on diverse tasks and dialogue datasets, permitting image splits into up to nine sub-images for processing higher resolution inputs up to 1344×1344 pixels, with all model parameters unfrozen for fine-tuning task-specific capabilities.

The video training tasks include video captioning at different granularities, open-domain question answering, and multiple-choice questions. All of these video tasks are designed to improve the model's ability of extracting information from long videos, answering questions across multiple frames, and performing spatio-temporal reasoning. To preserve image-understanding capabilities during training, we sample and combine data from the final sub-stage of image training with video data.

\subsection{Audio Alignment}
\label{subsec:audio_align}

\paragraph{Architecture}

The audio module of Capybara-OMNI is composed of an audio encoder and a MLP adapter. 
Inspired by the remarkable performance demonstrated by Qwen2-Audio~\citep{Qwen2-Audio} in audio understanding tasks, where its audio encoder is found on the Whisper-large-v3~\citep{radford2022whisper}, we initialize the audio encoder from Qwen2-Audio. 
The audio data is resampled at a frequency of 16 kHz. Moreover, a pooling layer with a stride of 2 is utilized to decrease the quantity of audio tokens. The audio encoder maps approximately one second long audio segments to around 25 tokens. Then, a single-layer MLP is applied to map the output of the speech encoder to the input space of the LLM, thereby facilitating the integration of audio signals into the LLM.

\paragraph{Training Data}

During the audio alignment stage, the audio module parameters are fine-tuned by using automatic speech recognition (ASR) and speech-to-text translation (S2TT) tasks. The source training data is presented in Table~\ref{tab:1}. In the Chinese ASR task, the training datasets predominantly consist of FreeST~\citep{FreeST}, Aishell~\citep{bu2017aishell1opensourcemandarinspeech}, WenetSpeech4TTS (Premium)~\citep{ma2024wenetspeech4tts}, and Zhvoice~\citep{zhvoice}. For the English ASR task, we use the LibriTTS~\citep{zen2019librittscorpusderivedlibrispeech} and VCTK~\citep{vctk} datasets. Notably, WenetSpeech4TTS is derived from the WenetSpeech dataset. It has undergone meticulous processing and is partitioned into subsets of varying scales based on the speech quality levels. Similarly, LibriTTS is derived from the LibriSpeech corpus. Through techniques such as noise elimination, text normalization, and correction of speech segmentation time points, the data quality has been markedly improved. For the S2TT task, the CoVoST2 dataset~\citep{wang2020covost} is utilized. To address the problem of insufficient Chinese-to-English translation data, we apply ASR and machine translation methods to sample and synthesize Chinese-to-English translation training data from CommonVoice~\citep{commonvoice:2020}.

\begin{table*}[!htbp]
\caption{Training data for Audio Alignment.}
\centering
\setlength{\tabcolsep}{4pt}
\renewcommand{\arraystretch}{1.2}
\resizebox{\textwidth}{!}{%
\begin{tabular}{@{}l|c|c|c|c@{}}
\toprule
{\textbf{Data Scenario}} & \textbf{Description} & \textbf{Dataset} & \textbf{Questions (K)} & \textbf{Language} \\ 
\midrule
\multirow{5}{*}{ASR} & \multirow{5}{*}{Automatic Speech Recognition} & WenetSpeech4TTS (Premium) & 400 & ZH \\
& & FreeST & 100 & ZH \\
& & Aishell & 110 & ZH \\
& & Zhvoice & 1130 & ZH \\
& & LibriTTS & 120 & ENG \\
& & VCTK & 40 & ENG \\
\cline{1-5}
S2TT & Speech-to-Text Translation & CoVoST2 & 100 & ZH\_ENG $\vert$ ENG\_ZH \\
\bottomrule
\end{tabular}%
}
\label{tab:1}
\end{table*}

Data filtering plays a crucial role in guaranteeing the quality and reliability of the training dataset. 
Thus, a meticulously curated ensemble of high-quality data screening methods for audio alignment has been assembled in our experiments.
In the context of ASR tasks, the data typically encompasses audio files and their corresponding texts. 
Initially, duplicates within the collected data are eliminated based on text content. Furthermore, with the aim of augmenting diversity while regulating the volume of training data, the training data is clustered according to semantics, and samples that are overly alike are filtered out. 
Subsequently, taking into account that some audio files might contain excessive noise or there could be incongruities between the audio and the text, open-source ASR tools are employed to predict the content of the audio files. 
The predictions are then compared to the corresponding ground-truth text, and samples with substantial discrepancies (Word Error Rate greater than 0.3) are filtered out to ensure data quality.
For the data of S2TT tasks, a parallel approach is adopted. 
Specifically, ASR tools are utilized to recognize the audio in the data, and then the Qwen2 model~\citep{qwen2} is used to translate the recognition results into the target language. Owing to the uncertainty inherent in the translation results, semantic similarity, rather than Word Error Rate, is adopted as the judgment criterion. Through data filtering, approximately 1.4 million data instances were obtained for the training during the audio alignment stage, which is less than the quantity of training data typically demanded by general audio-language models.

\paragraph{Training Process}

During the visual alignment stage, the model has established a robust foundation in image and video understanding. 
In the course of audio alignment training, it is essential not only to enable the LLM to comprehend audio inputs but also to prevent any negative impacts on its understanding capabilities of visual modalities. 
Consequently, we follow the Freeze-Omni~\citep{wang2024freeze} solution. 
Its primary contribution resides in addressing the challenge of how to avert catastrophic forgetting of the LLM under conditions of limited data and computational resources, while endowing it with the capacity to understand the speech modality. 
Specifically, throughout the training process, the parameters of the LLM and the visual component are kept frozen, and the audio encoder and adapter components are trained with a learning rate of 5e-5 for 5 epochs. 

\section{Experiments}
\label{sec:experiment}

Capybara-OMNI is trained upon a vision-language model and incorporates audio-understanding capabilities. Thus, in this section, the understanding capabilities of Capybara-OMNI in the image, video, and audio modalities are verified resectively. For the evaluation of each modality, open-source evaluation datasets and evaluation protocols are utilized to ensure that the results are as realistic and comparable as feasible. 
Subsequently, ablation experiments are carried out on the audio-understanding task to validate the effectiveness of two key measures: 1) Initializing the audio encoder with Qwen2-Audio; 2) Improving the audio alignment training data quality.

\paragraph{Image Understanding}

Capybara-OMNI shows highly competitive performance in a diverse array of tasks. 
It consistently attains either the top or second-place rankings on numerous benchmarks. 
The exceptional prowess of the astrophysics is noteworthy in scientific and mathematical reasoning. 
On certain benchmarks, the results achieved by Capybara-OMNI are not only on par with but often surpass those of open-source models that are an order of magnitude larger. 
In addition, it outperforms leading closed source models such as LLaVA-OV-72B and GPT4o-mini, highlighting its remarkable capabilities in these critical domains. The outcomes are presented in Table~\ref{tab:2}.

To evaluate the performance of Capybara-OMNI in real-world scenarios, we employ general question-answering and human-interaction tasks. Specifically, HallusionBench~\citep{guan2024hallusionbenchadvanceddiagnosticsuite} is utilized for hallucination detection, while MMVet~\citep{yu2023mmvetevaluatinglargemultimodal} and MMBench~\citep{liu2023mmbench} are adopted for comprehensive evaluation. Additionally, MathVista~\citep{lu2023mathvista} is used to assess mathematical reasoning capabilities, and MMStar~\citep{chen2024we} and MMMU~\citep{yue2023mmmu} are employed for evaluating general knowledge and reasoning. Finally, we assess the model's abilities in optical character recognition (OCR) and understanding text-rich images, with OCRBench~\citep{liu2024ocrbench} and AI2D~\citep{kembhavi2016diagram} being used for scientific diagram question-answering evaluation.

Capybara-OMNI showcases excellent performance across various scenarios, often leading the pack. On HallusionBench and MMBench leaderboards, it performs on par with or slightly better than Qwen2-VL-7B, dominating among comparable-scale open-source VLMs.
In the MMVet framework's free-form question-answering tasks, Capybara-OMNI ranks second, behind only Qwen2-VL-7B, and significantly outperforms MiniCPM-V2.6 (score 60.0) and InternVL2-8B (score 54.3), highlighting its strong conversational skills. Capybara-OMNI outshines contemporary models with similar parameter counts, achieving the highest average performance on MathVista and MMStar. On MathVista, it scores an average of 68.3, comparable to models with over ten-fold more parameters like InternVL2-76B (65.5) and LLaVA-OV-72B (67.5), demonstrating its enhanced multimodal reasoning. It also performs well in comprehensive and scientific reasoning tasks, validating its versatility. On OCRBench, it shows strong OCR capabilities, outperforming many similar-sized competitors, and on AI2D, it scores 84.3, surpassing all rivals.

\begin{table*}[t!]
\caption{
Performance of models on HallusionBench, MMVet, MMBench, MathVista, MMStar, MMU, OCRBench, and AI2D, OCRBench scores are normalized to a maximum of 100. Results are from official sources of the model or benchmark, or evaluated by use when unavailable. We highlight three groups: \colorbox{Gray!20}{private models}, open-source \colorbox[HTML]{d3f4ff}{vlms with one size larger},  open-source \colorbox[HTML]{e1f8bd}{vlms with comparable size}, and previous \colorbox[HTML]{ebdef0}{omni models} with both visual and audio capacity.}
\centering
\setlength{\tabcolsep}{4pt}
\renewcommand{\arraystretch}{1.2}
\resizebox{\textwidth}{!}{%
\begin{tabular}{@{}l|cccccccc|c}
\toprule
\textbf{Model} & \textbf{HallBench} & \textbf{MMVet} & \textbf{MMBench} & \textbf{MathVista} & \textbf{MMStar} & \textbf{MMU} & \textbf{OCRBench} & \textbf{AI2D} & \textbf{average} \\
\cmidrule(l){2-9}
& average & GPT4-turbo & V1.1 & mini & test & val & en & test & \\
\midrule
\rowcolor{Gray!20}
GPT4o~\citep{openai2024gpt4o} & 55.0 & 69.1 & 83.1 & 63.8 & 64.7 & 69.1 & 73.6 & 84.6 & 70.4 \\
\rowcolor{Gray!20}
Gemini1.5-pro~\citep{team2024gemini} & 45.6 & 64.0 & 74.6 & 63.9 & 59.1 & 62.2 & 75.4 & 79.1 & 65.5 \\
\rowcolor{Gray!20}
Claude3.5-sonnet~\citep{anthropic2024claude35} & 49.9 & 66.0 & 78.5 & 67.7 & 62.2 & 68.3 & 78.8 & 81.2 & 69.1 \\
\rowcolor{Gray!20}
GPT4V~\citep{openai2023gpt4v} & 43.9 & 67.5 & 79.8 & 58.1 & 56.0 & 63.1 & 64.5 & 78.2 & 63.9 \\
\rowcolor{Gray!20}
GPT4o-mini~\citep{openai2024gpt4o} & 46.1 & 66.9 & 76.0 & 52.4 & 54.8 & 60.0 & 78.5 & 77.8 & 64.1 \\

\midrule
\rowcolor{lightblue}
InternVL2-76B~\citep{chen2023internvl} & 55.2 & 64.4 & 85.5 & 65.5 & 67.4 & 62.7 & 83.9 & 87.6 & 71.5 \\
\rowcolor{lightblue}
Qwen2-VL-72B~\citep{wang2024qwen2vlenhancingvisionlanguagemodels} & 58.7 & 73.9 & 85.9 & 70.5 & 68.6 & 64.5 & 87.7 & 88.1 & 75.8 \\
\rowcolor{lightblue}
LLaVA-OV-72B~\citep{li2024llavaonevisioneasyvisualtask} & 47.9 & 60.6 & 84.5 & 67.5 & 65.8 & 56.8 & 74.1 & 85.6 & 68.1 \\

\midrule
\rowcolor{teagreen}
Molmo-7B-D~\citep{deitke2024molmopixmoopenweights} & 47.7 & 53.3 & 70.9 & 47.3 & 54.4 & 48.7 & 69.4 & 79.6 & 60.2 \\
\rowcolor{teagreen}
LLaVA-OV-7B~\citep{li2024llavaonevisioneasyvisualtask} & 31.6 & 51.9 & 80.9 & 63.2 & 61.7 & 48.8 & 62.2 & 81.4 & 60.2 \\
\rowcolor{teagreen}
MiniCPM-V2.6~\citep{yao2024minicpmvgpt4vlevelmllm} & 48.1 & 60.0 & 78.0 & 60.6 & 57.5 & 49.8 & 85.2 & 82.1 & 65.2 \\
\rowcolor{teagreen}
Intern-VL2-8B~\citep{chen2023internvl} & 45.2 & 54.3 & 79.4 & 58.3 & 62.0 & 52.6 & 79.4 & 83.8 & 64.4 \\
\rowcolor{teagreen}
Qwen2-VL-7B~\citep{wang2024qwen2vlenhancingvisionlanguagemodels} & 50.6 & 61.8 & 81.0 & 58.2 & 60.7 & 54.1 & 86.6 & 83.0 & 67.0 \\

\midrule
\rowcolor{lightpink}
VITA1.0~\citep{fu2024vitaopensourceinteractiveomni} & 39.7 & 41.6 & 71.8 & 44.9 & 46.4 & 47.3 & 67.8 & 73.1 & 54.1 \\
\rowcolor{lightpink}
VITA1.5-Audio~\citep{fu2025vita15gpt4olevelrealtime} & 44.9 & 49.6 & 76.7 & 66.2 & 59.9 & 52.1 & 73.2 & 79.3 & 62.7 \\

\midrule
\midrule
\rowcolor{front-color}
\textit{CAPYBARA-OMNI} (ours) & 50.6 & 60.5 & 81.3 & 68.3 & 63.9 & 52.3 & 84.4 & 84.3 & 68.2 \\
\bottomrule
\end{tabular}%
}

\label{tab:2}
\end{table*}

\paragraph{Video Understanding}

\begin{table*}[t!]
\caption{
Performance of models on video benchmarks. Results of other models are collected from official sources of the model or benchmark. We highlight three groups: \colorbox{Gray!20}{private models}, open-source \colorbox[HTML]{d3f4ff}{vlms with one size larger},  open-source \colorbox[HTML]{e1f8bd}{vlms with comparable size}, and previous \colorbox[HTML]{ebdef0}{omni models} with both visual and audio capacity.
}
\centering
\setlength{\tabcolsep}{4pt}
\renewcommand{\arraystretch}{1.2}
\resizebox{\textwidth}{!}{%
\begin{tabular}{@{}l|cccccccc}
\toprule

\multirow{2}{*}{\textbf{Model}} &  \multicolumn{3}{c}{\textbf{Video-MME}}    & \textbf{MVbench} &  \textbf{PerceptionTest} & \textbf{ActivityNETQA} & \textbf{DREAM-1K} \\
& frames & w/o subtitles & with subtitles & & &    &  \\
\midrule
\rowcolor{Gray!20}
GPT4o~\citep{openai2024gpt4o} & 384 & 71.9 & 77.2 & - &  - & - & 39.2 \\
\rowcolor{Gray!20}
Gemini1.5-pro~\citep{team2024gemini} & 1/0.5 fps & 75.0 & 81.3 & 81.3  & - & 57.5 & 36.2 \\
\rowcolor{Gray!20}
Claude3.5-sonnet~\citep{anthropic2024claude35} & 20 & 60.0 & 62.9 & - & - & - & 31.2 \\
\rowcolor{Gray!20}
GPT4V~\citep{openai2023gpt4v} & 10 & 59.9 & 63.3 & 43.7& - & 57.0 & 34.4 \\
\rowcolor{Gray!20}
GPT4o-mini~\citep{openai2024gpt4o} & 250 & 64.8 & 68.9 &  - & - & - & 34.0 \\

\midrule

\rowcolor{lightblue}
InternVL2-76B~\citep{chen2023internvl} & 16 & 61.2 &  - & - & - & - & 31.7 \\
\rowcolor{lightblue}
Qwen2-VL-72B~\citep{wang2024qwen2vlenhancingvisionlanguagemodels} & 2 fps, 768 & 71.2 & 77.8 & 73.6 & 68.0 & - & 33.2 \\
\rowcolor{lightblue}
LLaVA-OV-72B~\citep{li2024llavaonevisioneasyvisualtask} & 32 & 66.3 & 69.6 & 59.4 & 57.1 & 62.3 & 33.2 \\
\rowcolor{lightblue}
LLaVA-Video-72B \citep{zhang2024videoinstructiontuningsynthetic} & 64 & 70.5 & 76.9 & 74.4 & 74.3 & 63.4 & 34 \\


\midrule
\rowcolor{teagreen}
\rowcolor{teagreen}
LLaVA-Video-7B \citep{zhang2024videoinstructiontuningsynthetic} & 64 & 63.3 & 69.7 & 58.6 & 67.9 & 56.5 & 32.5 \\
\rowcolor{teagreen}
LLaVA-OV-7B~\citep{li2024llavaonevisioneasyvisualtask} & 64 & 58.2 & 60.1 & 56.7 & 57.1 & 56.6 & 31.7 \\
\rowcolor{teagreen}
MiniCPM-V2.6~\citep{yao2024minicpmvgpt4vlevelmllm} & 64 & 60.9 & 63.7 & -  & - & - & 30.5 \\
\rowcolor{teagreen}
\rowcolor{teagreen}
Intern-VL2.5-8B  & 64 & 64.2 & 66.9 & 72.0& - & - & -\\
\rowcolor{teagreen}
Qwen2-VL-7B~\citep{wang2024qwen2vlenhancingvisionlanguagemodels} & - & 63.3 & - & 67.0 & 62.3 & - & 29.6 \\
\midrule

\rowcolor{lightpink}
VITA1.0~\citep{fu2024vitaopensourceinteractiveomni} &  - & 55.8 & 59.2 & - & - & - & 28.9 \\
\rowcolor{lightpink}
VITA1.5-Audio~\citep{fu2025vita15gpt4olevelrealtime} & - & 56.8 &  59.5 & 56.8 &  - & - & - \\
\midrule
\midrule

\rowcolor{front-color}
\textit{CAPYBARA-OMNI} (ours) & 1 fps, 128  & 65.7 & 70.0 &  67.8 & 74.8  & 67.6 & 35.2  \\
\bottomrule
\end{tabular}%
}

\label{tab:3}
\end{table*}

The video understanding ability is evaluated using two main categories of benchmarks: a set of video understanding and comprehension tasks, including Video-MME\citep{fu2024video}, MVBench\citep{li2023mvbench}, and PerceptionTest\citep{patraucean2023perception}, which focus on varying durations and cognitive domains. Additionally, Dream-1k\citep{wang2024tarsierrecipestrainingevaluating} is used for assessing captioning performance on complex video sequences. 

The evaluation results are listed in Table \ref{tab:3}. Firstly, Capybara-Omni demonstrates superior performance compared to other Vision-Language Models (VLMs) of similar sizes. For example, on the comprehensive benchmark set, Video-MME, our model outperforms both LLaVA-Video-7B and the recently released Intern-VL2.5-8B.
Secondly, Capybara-OMNI surpasses VITA1.5, the current state-of-the-art open-source omni model, by a significant margin.
These results highlight the effectiveness of the video understanding data engineering and training strategies. Furthermore, thanks to the training strategies designed to mitigate catastrophic forgetting, Capybara-Omni excels in retaining its video understanding capabilities.

\paragraph{Audio Understanding}

Table~\ref{tab:4} depicts the performance of Capybara-OMNI on the audio leaderboards. For Chinese ASR tasks, the Character Error Rate (CER) is adopted as the evaluation metric. In the case of English ASR tasks, the Word Error Rate (WER) serves as the evaluation metric, and for S2TT tasks, the BLEU score is employed as the evaluation metric. Aishell-1 is a commonly utilized Chinese ASR test set. As presented in Table~\ref{tab:4}, Capybara-OMNI (2.1) outperforms a diverse range of models and even marginally exceeds the specialized audio large-language model GLM-4-Voice (2.5). In the English ASR task on the LibriSpeech dataset, Capybara-OMNI also exhibits strong competitiveness. It is noteworthy that although the Audio Encoder module of Qwen2-Audio was utilized for the initialization of the Audio Encoder in our model, our model performs slightly less favorably than Qwen2-Audio in both ASR and S2TT tasks. This can likely be attributed to the fact that during the audio alignment process, the parameters of the LLM were not updated. This also represents a direction worthy of future exploration.

\begin{table*}[t!]
\caption{
Performance of models on Audio BenchMark. Aishell-1 and FLEURS are Chinese test sets for the ASR task, where we evaluate the performance using the CER metric. LibriSpeech is an English test set for the ASR task, evaluated using the WER metric. We highlight two groups of models: \colorbox[HTML]{d3f4ff}{audio-language models}, and \colorbox[HTML]{e1f8bd}{omni models} with comparable size.
}
\centering
\setlength{\tabcolsep}{4pt}
\renewcommand{\arraystretch}{1.2}
\resizebox{\textwidth}{!}{%
\begin{tabular}{@{}l|cccccccc}
\toprule
\multirow{2}{*}{\textbf{Model}} & \textbf{Aishell-1(↓) } & \textbf{FLEURS(↓) } & \textbf{LibriSpeech(↓) } & \textbf{LibriSpeech(↓) } & \textbf{LibriSpeech(↓) } & \textbf{LibriSpeech(↓) } & \textbf{S2TT(↑) } \\ 
& test(zh) & test(zh) & dev-clean(en) & dev-other(en) & test-clean(en) & test-other(en) & en\_zh $\vert$ zh\_en \\ 
\midrule

\rowcolor{lightblue}
Qwen2-Audio~\citep{Qwen2-Audio}& - & 7.1 & 1.3 & 3.4 & 1.6 & 3.6 &  45.2 $\vert$ 24.4\\ 
\rowcolor{lightblue}
GLM-4-Voice~\citep{zeng2024glm4voiceintelligenthumanlikeendtoend} & 2.5 & - & -  &  -  & 2.8 & 7.6 & -\\ 

\midrule 
\rowcolor{teagreen}
Mini-Omni2~\citep{xie2024mini2} & - & - & 4.8 & 9.8 & 4.7 & 9.4 & -\\ 
\rowcolor{teagreen}
Freeze-Omni ~\citep{wang2024freeze}& 2.8 & - & 4.2 & 10.2 & 4.1 & 10.5 & -\\ 
\rowcolor{teagreen}
VITA-1.0 ~\citep{fu2024vitaopensourceinteractiveomni}& - & - & 7.6 & 16.6 & 8.1 & 18.4 & -\\
\rowcolor{teagreen}
VITA-1.5 ~\citep{fu2025vita15gpt4olevelrealtime}& 2.2 & - & 3.3 & 7.2 & 3.4 & 7.5 & -\\ 
\midrule
\midrule

\rowcolor{front-color}
\textit{CAPYBARA-OMNI} (ours) & 2.1 & 7.4 & 2.4 & 5.6 & 2.3 & 5.2 & 41.6 $\vert$ 23.6\\ 
\bottomrule
\end{tabular}%
}

\label{tab:4}
\end{table*}

\paragraph{Ablation Study}

In this chapter, an investigation is conducted into the impact of employing different audio encoders and data-quality enhancement methods on the model's audio-understanding ability. The model with ``whisper-large + one-layer MLP'' as the audio encoder and without any data-quality enhancement served as the baseline. As presented in Table~\ref{tab:5}, we discover that, after initialization with the audio encoder of Qwen2-Audio, the model achieves substantial improvements in the two tasks. This can be ascribed to the fact that Qwen2-Audio was pretrained using a large quantity of high-quality data, thereby injecting copious audio information into the audio encoder. Moreover, subsequent to enhancing the data quality, the model's audio capabilities are further enhanced. The primary reasons are that the negative impact of noise on training is circumvented, and simultaneously, the data distribution becomes more rational, covering a more diverse range of audio scenarios.

\begin{table*}[t!]
\caption{
Comparative study of different audio encoders and training data screening strategies.
}
\centering
\setlength{\tabcolsep}{4pt}
\renewcommand{\arraystretch}{1.2}
\resizebox{\textwidth}{!}{%
\begin{tabular}{@{}l|cccccccc}
\toprule
\multirow{2}{*}{\textbf{Model}} & \textbf{Aishell-1(↓) } & \textbf{FLEURS(↓) } & \textbf{LibriSpeech(↓) } & \textbf{LibriSpeech(↓) } & \textbf{LibriSpeech(↓) } & \textbf{LibriSpeech(↓) } & \textbf{S2TT(↑) } \\ 
& test(zh) & test(zh) & dev-clean(en) & dev-other(en) & test-clean(en) & test-other(en) & en\_zh $\vert$ zh\_en \\ 
\midrule 
\rowcolor{teagreen}
BASELINE & 7.4 & 9.7 & 6.4 & 6.7 & 4.7 & 7.6 & -\\ 
\midrule
\rowcolor{teagreen}
+Qwen2-Audio Encoder& 2.6 & 7.9 & 3.5 & 6.3 & 3.1 & 5.8 & 40.9 $\vert$ 22.8 \\ 
\midrule
\rowcolor{teagreen}
+Data augmentation and improvement & 2.1 & 7.4 & 2.4 & 5.6 & 2.3 & 5.2 & 41.6 $\vert$ 23.6\\ 
\bottomrule
\end{tabular}
}

\label{tab:5}
\end{table*}
\section{Cross-Modal Instruction Tuning}
The previous work has imbued the Capybara-OMNI with robust visual and audio comprehension capabilities. Nevertheless, it still falls short in cross-modal data understanding and human interaction aspects. Consequently, we introduce the cross-modal instruction tuning phase. This phase encompasses carefully crafted training data incorporating multiple modalities, along with chat data formulated in conjunction with video content, endowing the model with favorable interaction capabilities and formidable cross-modal understanding capabilities.

\paragraph{Training Data}
First, we use prompts to guide GPT-4o~\citep{openai2024gpt4o} to understand video frames, and then construct alternating dialogue data by integrating with the video content. The generated question-answer pairs should simulate the state of casual conversations on a daily basis as much as possible. The prompts used are provided in Figure~\ref{fig:4} and Figure~\ref{fig:5} in the Appendix.
This approach simulates the interaction scenarios between the omni model and humans during its usage. The generation results are shown in Figure~\ref{fig:6} in the Appendix.
In addition, we also extracted some Visual Question Answering (VQA) data from the previous training data. This part of the data helps activate the visual comprehension ability acquired during the training in the previous stage. 
We converted the text parts of the data from the two parts into audio through Text-to-Speech (TTS) technology for the text-formatted questions, thereby constructing cross-modal training data in the form of video + audio, and achieving the alignment between audio and text.
Ultimately, we obtained approximately 100K pieces of cross-modal fine-tuning data conducive to activating the visual comprehension ability acquired during the historical training phase, and simultaneously achieves the alignment between audio and text.

\paragraph{Training Process}
As shown in Figure~\ref{fig:2}, in order to achieve better results during the cross-modal instruction tuning stage, all the parameters in the model are activated and updated. We set the learning rate to 1e-5 and train the model for 3 epochs to obtain the optimal performance. The training process is extremely efficient that takes around 3 hours. In addition, referring to the deployment framework of VITA, we build an interactive demo for the model (See the detailed demo in the supplementary materials).
\section{Conclusions}
In this work, we introduce Capybara-OMNI that presents an efficient and lightweight paradigm for building omni-modal language models.
Through a carefully constructed training strategy and training data, the model demonstrates remarkable comprehension abilities in image understanding, video understanding, and audio understanding. 
In particular, the extension of full-modality can be achieved through low complexity and few resources.
Numerous evaluation experiments show that Capybara-OMNI achieves remarkable performance in various multimodal benchmarks. 
In the part of audio alignment, by optimizing the training strategy, while resolving the inherent conflicts among data of different modalities, the model can be endowed with powerful audio comprehension capabilities merely by using a small amount of open-source data. 
Through cross-modal instruction tuning, the model achieves cross-modal understanding and speech-based interaction. 
We hope that Capybara-OMNI can continue to promote the development of open-source models in the field of multimodal interaction.

\normalem
\bibliography{iclr2025_conference}
\bibliographystyle{iclr2025_conference}

\appendix
\section{Appendix}

\begin{table*}[htbp]
\centering
\resizebox{\textwidth}{!}{
    \begin{tabular}{l|l}
    \toprule
    Task & Dataset \\
    \midrule
    
    \multirow{3}{*}{Caption} 
    & 
    Allava-Caption~\citep{chen2024allava}, IIW~\citep{garg2024imageinwords}, ShareGPT4V~\citep{chen2023sharegpt4v}, \\
    &
    ShareGPT4o~\citep{sharegpt4o}, TextCaps~\citep{textocr-gpt4v}, Image Textualization~\citep{pi2024image}, \\
    & 
    DOCCI~\citep{OnoeDocci2024}, TextOCR-GPT4V~\citep{textocr-gpt4v}, PVIT~\citep{chen2023positionenhanced}
    \\
    \midrule

    \multirow{6}{*}{General QA}
    &
    VQAv2~\citep{antol2015vqa}, GQA~\citep{hudson2019gqa}, VisDial~\citep{das2017visualdialog}, \\
    &  
    IconQA~\citep{lu2021iconqa}, TallyQA~\citep{acharya2019tallyqa}, VizWiz~\citep{gurari2018vizwiz}, \\
    & 
    RefCOCO~\citep{yu2016modelingcontextreferringexpressions}, IconQA~\citep{lu2021iconqa}, CLEVR~\citep{johnson2017clevr}, \\
    & 
    Hateful Memes~\citep{kiela2020hateful}, IDK~\citep{cha2024visuallydehallucinativeinstructiongeneration},
    SketchyVQA~\citep{tu2023many}, \\
    &
    VSR~\citep{Liu2022VisualSR}, Visual7W~\citep{zhu2016visual7w}, ComVint~\citep{du2023makes}, \\
    &
    Vision FLAN~\citep{xu2024vision}, MMInstruct~\citep{liu2024mminstruct}, LRV Normal~\citep{liu2023aligning}

    \\
    \midrule

    \multirow{3}{*}{Conversations}
    &  
    ALLaVA Instruct~\citep{chen2024allava}, LLaVA-158K~\citep{liu2023llava}, LLaVAR~\citep{zhang2023llavar}, \\
    & 
    Cambrian-GPT4v~\citep{tong2024cambrian}, Cambrian-GPT4o~\citep{tong2024cambrian}, SVIT~\citep{zhao2023svitscalingvisualinstruction}, \\
    & 
    ShareGPT4v-Dailogue~\citep{chen2023sharegpt4v}, RLAIF-V~\citep{yu2024rlaifvopensourceaifeedback}
    \\
    \midrule

    \multirow{6}{*}{OCR \& QA }
    &
    ORCAND-CAR~\citep{zhan2017handwrittendigitstringrecognition},
    HWL-en~\citep{tal2023hwl},
    ChromeWriting~\citep{RenderedText}, \\
    &
    HME100K~\citep{yuan2022syntax},
    IAM~\citep{marti2002iam},
    RCTW-17~\citep{shi2017icdar2017}, \\
    &
    HierText~\citep{long2023icdar}, TextOCR~\citep{singh2021textocr}, ReCTS~\citep{liu2019icdar}, \\
    &
    LSVT~\citep{sun2019icdar}, Rendered Text~\citep{RenderedText}, SynthDog-EN~\citep{kim2022donut}, \\
    &
    SynthDog-ZH~\citep{kim2022donut}, 
    TextVQA~\citep{singh2019towards},
    ST-VQA~\citep{biten2019scene},\\

    & 
    EST-VQA~\citep{wang2020general} 
    DT-VQA~\citep{zhang2024exploring}

    \\
    \midrule

    \multirow{5}{*}{Doc \& KIE }
    
    &
    idl-wds~\citep{idl-ws},
    pdfa-wds~\citep{pdfa-eng-wds},
    DocStruct4M~\citep{hu2024mplugdocowl15unifiedstructure}, \\
        
    &
    RobutSQA~\citep{zhao2023robutsystematicstudytable},
    RobutWiki~\citep{zhao2023robutsystematicstudytable}, 
    RobutWTQ~\citep{zhao2023robutsystematicstudytable}, \\
    
    &
    UReader IE~\citep{ye2023ureader},
    UReader KG~\citep{ye2023ureader},
    UReader QA~\citep{ye2023ureader}, \\

    &
    VisualMRC~\citep{tanaka2021visualmrc},
    InfoVQA~\citep{mathew2022infographicvqa},
    ScreenQA~\citep{hsiao2022screenqa}, \\

    & 
    HiTab~\citep{cheng2021hitab},
    MMTab~\citep{zheng2024multimodaltableunderstanding}
    Screen2Words~\citep{wang2021screen2words}, \\
  
    & 
    DocVQA~\citep{mathew2021docvqa},
    DocLayNet~\citep{pfitzmann2022doclaynet},
    OCR-VQA~\citep{mishraICDAR19},\\
    
    &
    Docmatix~\citep{laurenccon2024building}, DocReason25K~\citep{hu2024mplugdocowl15unifiedstructure},
    FUNSD~\citep{jaume2019funsddatasetformunderstanding}, \\
   
    &   
    SROIE~\citep{Huang_2019}, POIE~\citep{kuang2023visual},  WildReceipt~\citep{sun2021spatial}\\

    \midrule

    \multirow{5}{*}{Chart}
    & 
    PlotQA~\citep{methani2020plotqa},
    DVQA~\citep{kafle2018dvqa},
    FigureQA~\citep{kahou2018figureqaannotatedfiguredataset}, \\

    &
    ArxivQA~\citep{li2024multimodal},
    VisText~\citep{tang2023vistext},
    OpenCQA~\citep{kantharaj2022opencqa}, \\

    &
    Chart2Text~\citep{obeid2020chart},
    LRV Chart~\citep{liu2023aligning},
    TinyChart~\citep{zhang2024tinychartefficientchartunderstanding},\\

    &
    ChartQA~\citep{masry2022chartqa},
    ChartSumm~\citep{rahman2023chartsumm}, 
    MMC~\citep{liu2023mmc}, \\
    
    &
    ReachQA~\citep{he2024distillvisualchartreasoning}, ChartGemma~\citep{masry2407chartgemma}
    ChartLlama~\citep{han2023chartllamamultimodalllmchart}

    \\
    \midrule

    \multirow{7}{*}{Math}
    
    & 
    MAVIS Manual Collection~\citep{zhang2024mavismathematicalvisualinstruction},
    MAVIS Data Engine~\citep{zhang2024mavismathematicalvisualinstruction}, \\ 
    &
    
    UniGeo~\citep{chen2022unigeounifyinggeometrylogical} ,
    Geo170K Align~\citep{gao2023gllavasolvinggeometricproblem} ,
    Geo170K QA~\citep{gao2023gllavasolvinggeometricproblem}, \\ 
    &
    Geometry3K~\citep{lu2021intergpsinterpretablegeometryproblem},
    GEOS~\citep{seo2015solving},
    UniGeo (MathV360K)~\citep{kazemi2023geomverse},\\
    &
    GeoMVerse (MathV360K)~\citep{kazemi2023geomverse},
    GeoQA+ (MathV360K)~\citep{chen2022geoqageometricquestionanswering},  \\ 

    & MapQA (MathV360K)~\citep{chang2022mapqadatasetquestionanswering} ,  
    Geometry3K (MathV360K)~\citep{lu2021inter},\\  

    &
    MathQA~\citep{amini2019mathqainterpretablemathword} ,
    RAVEN~\citep{zhang2019raven},
    Super-CLEVR~\citep{li2023super}, \\ 
    &
    TabMWP~\citep{lu2023dynamic} ,
    InterGPS~\citep{lu2021intergpsinterpretablegeometryproblem},
    CLEVR-Math~\citep{johnson2017clevr},
    \\ \midrule

    \multirow{3}{*}{Knowledge}
    &
    KVQA~\citep{marino2019okvqa}, OK-VQA~\citep{marino2019okvqa}, A-OKVQA~\citep{schwenk2022aokvqa}, \\
    & 
    AI2D~\citep{kembhavi2016diagram}, 
    Cambrian DataEngine~\citep{tong2024cambrian},
    ScienceQA~\citep{lu2022learn} \\
    
    &
    TQA~\citep{kembhavi2017you}, WIT~\citep{srinivasan2021wit},
    ViQuAE~\citep{lerner2022viquae}

    \\
    \midrule

    Domains & 
    VQA-Rad~\citep{lau2018dataset}, 
    PMC-VQA~\citep{zhang2024pmcvqavisualinstructiontuning},
    WebSight~\citep{laurençon2024unlocking} ,  \\
    
    &
    CoInstruct~\citep{wu2024towards}
    knoiq-10k~\citep{Hosu_2020}

    \\
    \midrule
    Text & 
    Magpie Pro~\citep{xu2024magpie},
    Magpie Pro (L3 ST),
    Magpie Pro (Qwen2 ST)\\

    &
    AutoMathText~\citep{zhang2024automathtext},
    JiuZhang~\citep{zhou2024jiuzhang3}
    
    \\
    \bottomrule
    \end{tabular}
    }
    \caption{The training data used in the visual alignment process. In the first stage, we use the same 556k dataset as LLava. For the second stage, we use 80\% of the samples from the Caption, OCR \& QA, Doc \& KIE, and Chart categories, while the remaining 20\% and samples from other sources are reserved for the third stage. Datasets excluded during the source filtering stage are not included in the table.}
    \label{tab:6}
\end{table*}

\noindent
\begin{figure*}[!h]
\centering
\begin{tcolorbox}[colback=gray!10!white, colframe=black]
First, please act as a video chat user who is engaged in a video call with a friend. The input frames represent the video content on your side (i.e., the video information you are showing to your friend). Based on the specific content of the video, ask your friend questions from a first-person perspective.
Then, act as the other party and respond to the user's question based on the video content and the question asked. Alternate roles accordingly and create a dialogue consisting of three to five rounds.
Requirements:
1.Both the question and the response must be generated in Chinese.
2.Make the generated questions and answers as conversational as possible.
3.When playing the role of the friend, the responses can be humorous and witty.
4.Please output the final result in the following format: [{'quesiton':User's question,'answer':generated response},...,{'quesiton':User's question,'answer':generated response}].
\end{tcolorbox}
\caption{Prompt for generating single-round dialogues based on video content.}
\label{fig:4}
\end{figure*}

\begin{figure*}[!h]
\centering
\begin{tcolorbox}[colback=gray!10!white, colframe=black]
First, please act as a video chat user who is engaged in a video call with a friend. The input frames represent the video content on your side (i.e., the video information you are showing to your friend). Based on the specific content of the video, ask your friend questions from a first-person perspective.
Then, act as the other party and respond to the user's question based on the video content and the question asked. Alternate roles accordingly and create a dialogue consisting of three to five rounds.
Requirements:
1.Both the question and the response must be generated in Chinese.
2.Make the generated questions and answers as conversational as possible.
3.When playing the role of the friend, the responses can be humorous and witty.
4.The generated multi-turn dialogue should ideally maintain continuity between turns.
5.Please output the final result in the following format: [{'quesiton':User's question,'answer':generated response},...,{'quesiton':User's question,'answer':generated response}].
\end{tcolorbox}
\caption{Prompt for generating multi-round dialogues based on video content.}
\label{fig:5}
\end{figure*}

\begin{figure}[h]
\centering
\includegraphics[width=0.75\textwidth]{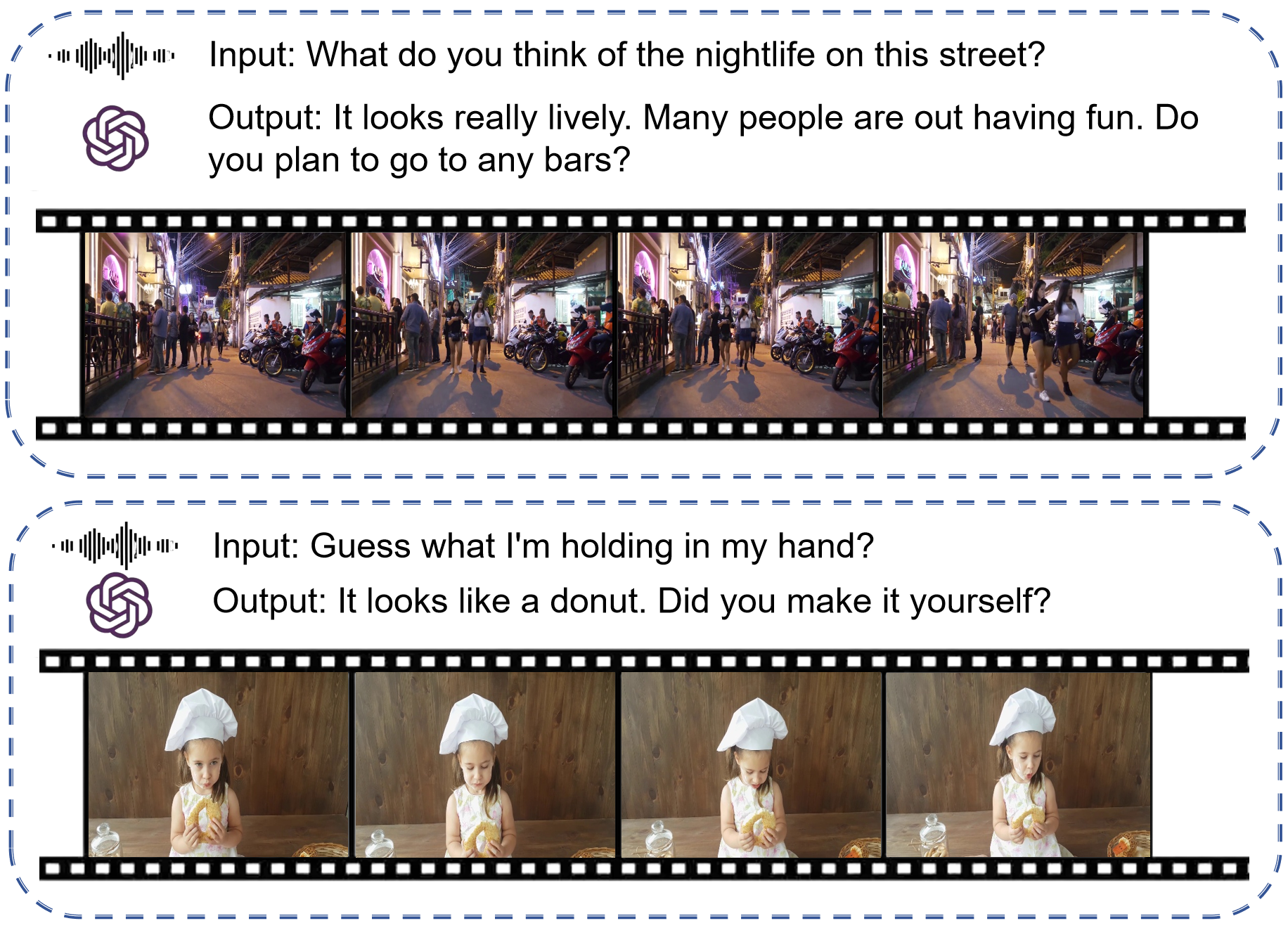}
\caption{example of cross-modal multimodal instruction data.}
\label{fig:6}
\end{figure}

\end{document}